\documentclass[lettersize,journal]{IEEEtran}
\usepackage{amsmath,amsfonts}
\usepackage{algorithmic}
\usepackage{algorithm}
\usepackage{array}
\usepackage{textcomp}
\usepackage{stfloats}
\usepackage{url}
\usepackage{verbatim}
\usepackage{graphicx}
\usepackage{cite}
\usepackage{subfigure}
\usepackage{multirow}
\usepackage{color}
\usepackage{booktabs}
\begin{document}

\newcommand{\model}{TransPrompt v2}

\title{\emph{\model}: A Transferable Prompting Framework for Cross-task Text Classification}


\author{Jianing Wang, Chengyu Wang, Cen Chen, Ming Gao, Jun Huang, Aoying Zhou

\thanks{Jianing Wang, Cen Chen, Ming Gao and Aoying Zhou with School of Data Science and Engineering, East China Normal University, Shanghai, 200062 China. E-mails: lygwjn@gmail.com, \{cenchen,mgao,ayzhou\}@dase.ecnu.edu.cn.
(Jianing Wang and Chengyu Wang contribute equally to the manuscript. Corresponding authors: Cen Chen and Ming Gao)}

\thanks{Chengyu Wang and Jun Huang are with Alibaba Group, Hangzhou, Zhejiang, 311121 China. E-mails: \{chengyu.wcy, huangjun.hj\}@alibaba-inc.com.}

\thanks{
This work was partially supported by Alibaba Group through the Alibaba Innovative Research Program and the National Natural Science Foundation of China under Grant Nos.
U1911203, 61877018, 61672234, and 61672384.
}}




\maketitle

\begin{abstract}

Text classification is one of the most imperative tasks in natural language processing (NLP). Recent advances with pre-trained language models (PLMs) have shown remarkable success on this task. However, the satisfying results obtained by PLMs heavily depend on the large amounts of task-specific labeled data, which may not be feasible in many application scenarios due to data access and privacy constraints.
The recently-proposed prompt-based fine-tuning paradigm improves the performance of PLMs for few-shot text classification with task-specific templates.
Yet, it is unclear how the prompting knowledge can be transferred across tasks, for the purpose of mutual reinforcement. 
We propose~\emph{\model}, a novel transferable prompting framework for few-shot learning across similar or distant text classification tasks. For learning across similar tasks, we employ a multi-task meta-knowledge acquisition (MMA) procedure to train a meta-learner that captures the cross-task transferable knowledge. For learning across distant tasks, we further inject the task type descriptions into the prompt, and capture the intra-type and inter-type prompt embeddings among multiple distant tasks. Additionally, two de-biasing techniques are further designed to make the trained meta-learner more task-agnostic and unbiased towards any tasks.
After that, the meta-learner can be adapted to each specific task with better parameters initialization. Extensive experiments show that~\emph{\model} outperforms single-task and cross-task strong baselines over multiple NLP tasks and datasets. We further show that the meta-learner can effectively improve the performance of PLMs on previously unseen tasks. In addition, \emph{\model} also outperforms strong fine-tuning baselines when learning with full training sets.
\footnote{An earlier version~\cite{wang2021transprompt} of this paper was presented at the 2021 Conference on Empirical Methods in Natural Language Processing (EMNLP 2021).}
\end{abstract}


\section{Introduction}

\IEEEPARstart{T}{ext} classification aims to classify a given text into a pre-defined class set~\cite{Minaee2021Deep}, which is a fundamental task and plays an important role in some natural language processing (NLP) scenarios, such as sentiment analysis~\cite{Zhang2020Knowledge, Bai2021Investigating}, natural language inference~\cite{Feng2022Neuro} and paraphrasing~\cite{Zulqarnain2021A}, etc.
Pre-trained Language Models (PLMs) have become the dominant infrastructures for text classification by employing the standard two-stage training, i.e., \emph{pre-training} and \emph{fine-tuning}. 
Notable PLMs include BERT~\cite{Devlin2019BERT}, RoBERTa~\cite{Liu2019RoBERTa}
GPT~\cite{radford2019language} and many others. 
Specifically, in the \emph{pre-training} stage, these PLMs follow the self-supervised training objective of Masked Language Modeling (MLM) or Auto-regressive Language Modeling (ALM) over large-scale unlabeled corpora to capture the intrinsic semantics and contextual information of the sentence. 
While in the \emph{fine-tuning} stage,
a majority of previous approaches utilize these well-designed PLMs with some new initialized classification heads to fine-tune on the task-specific data~\cite{Ma2021Deformable, Zhao2022EICO, Devlin2019BERT}.
Yet, these conventional methods heavily depend on the time-consuming and labor-intensive process of data annotation, which may be bothersome in real-world scenarios~\cite{Mukherjee2020Uncertainty, Gao2021Making, Wang2022Towards, liu2021ptuningv2, Hu2021Knowledgeable, Gu2021PPT}. 
Additionally, there is a large gap between the pre-training objective of MLM (i.e.,~the prediction distribution over the entire vocabularies) and the fine-tuning objective of the classification head (i.e.,~the prediction distribution of pre-defined classes), which hinders the transfer and adaptation of knowledge in PLMs to downstream tasks, and causes the over-fitting problem in low-resource scenarios~\cite{Schick2021Exploiting, Gao2021Making}.



\begin{figure}
\centering
\includegraphics[width=0.95\linewidth]{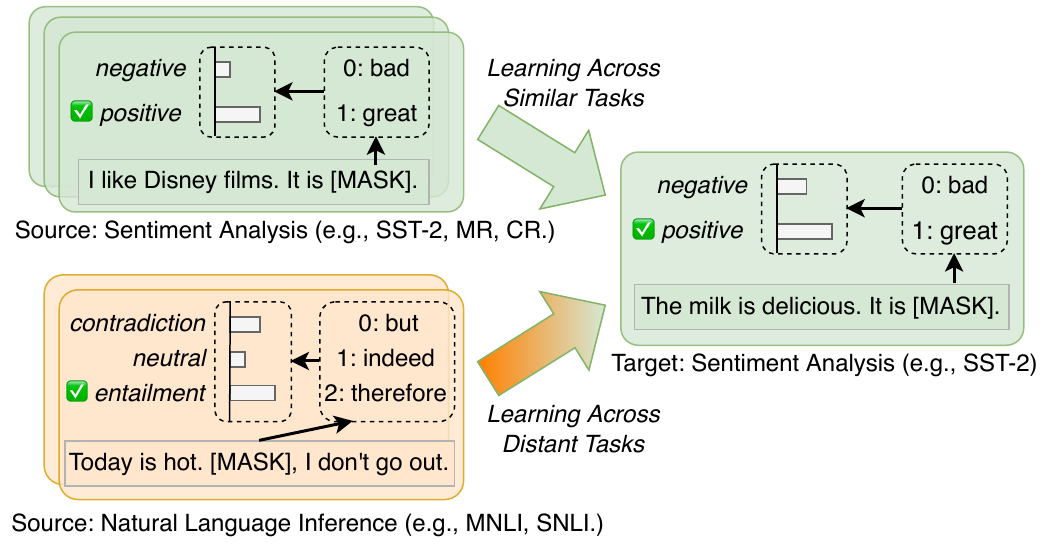}
\caption{Example of prompt-based learning across similar and distant tasks. The sentiment analysis task SST-2~\cite{Socher2013Socher} can benefit from similar tasks (e.g., SST-2, MR~\cite{Hu2004Mining} and CR~\cite{Pang2005Seeing}), and also from some distant tasks (e.g., MNLI~\cite{Williams2018A} and SNLI~\cite{Bowman2015A}).}
\label{fig:example}
\end{figure}

Recently, a branch of prompt-based learning (i.e., prompt-tuning or prompting) paradigm has arisen to transform the downstream classification task to the cloze-style problem, which makes the PLM easily adapt to specific tasks with very few labeled data. 
Specifically, the mature ultra-large PLM model GPT-3~\cite{Brown2020Language} presents in-context learning paradigm with some discrete templates in low-resource settings. 
Most recent methods~\cite{ Schick2021Exploiting, Gao2021Making, Min2022Noisy, Wu2022Adversarial, Hu2021Knowledgeable, Wang2022Towards} tune the BERT-style PLM over text classification tasks with well-designed task-specific discrete or continuous templates, which can support the reuse of the MLM objective. 
Take Figure~\ref{fig:example} as an example, in sentiment analysis, a prompt template with a masked language token~(e.g., ``It was \texttt{[MASK]}.'') is added to the review text (e.g., ``I like Disney films.''). We can reuse the vanilla MLM to obtain the result tokens of masked position for label prediction (e.g., ``great'' for the positive label and ``bad'' for the negative label).
Despite the remarkable success, we notice that current prompt-based approaches may have a few limitations. 
For few-shot learning, the performance of downstream tasks is still constrained by the number of training instances from a single task.
It would be highly desirable if the model can acquire the transferable knowledge from other tasks before it is adapted to specific tasks with few samples.
Therefore, a natural question arises:~\emph{how can we design a prompting framework for BERT-style models that captures transferable knowledge across different classification tasks to improve the performance of few-shot learning}?



\begin{figure}
\centering
\includegraphics[width=0.5\textwidth]{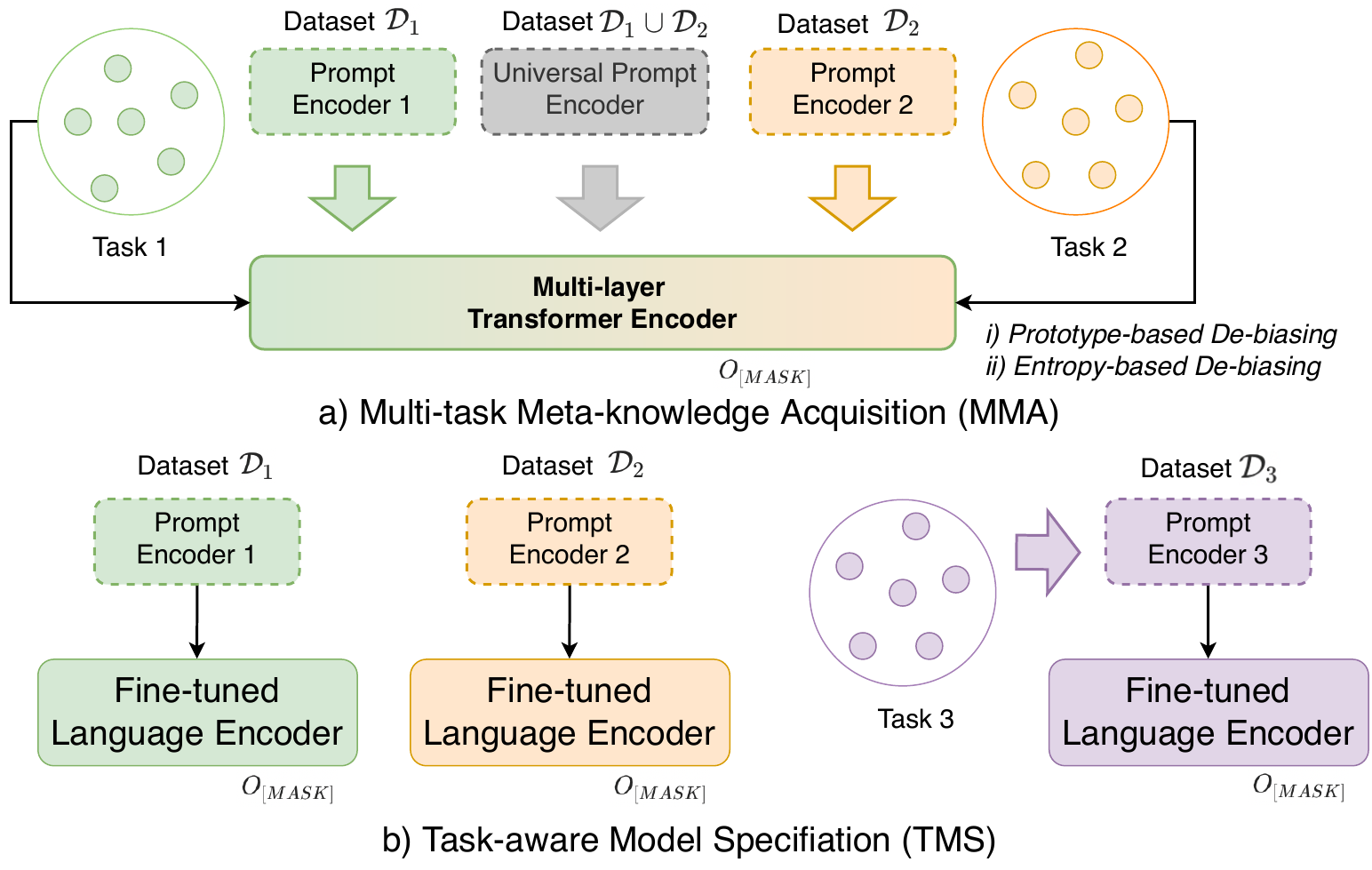}
\caption{The high-level architecture of the~\emph{\model} framework. In the toy example, Task 1 and Task 2 are existing tasks, while Task 3 is a new task for the meta-learner to generalize.  (Best viewed in color.)
}
\label{fig:overview}
\end{figure}

In this paper, we extend our previous work~\cite{wang2021transprompt} and present~\emph{\model}, a general prompting framework that allows PLMs to capture~\emph{cross-task transferable knowledge} for few-shot text classification.
Specifically, to improve the application scope of our algorithm,~\emph{\model} considers two scenarios shown in Figure~\ref{fig:example}: i) learning across~\emph{similar tasks}; and ii) learning across~\emph{distant tasks}.
For example, when we aim at training a sentiment classifier over the target task SST-2~\cite{Socher2013Socher}, we can directly obtain some similar tasks as the source data,
such as SST-2, MR~\cite{Hu2004Mining} and CR~\cite{Pang2005Seeing}. In contrast, when similar datasets are unavailable for us (which is a more challenging setting), the model can also be improved by leveraging some training data from distant source tasks~\footnote{Generally, such distant tasks have very different label spaces and contextual semantics from source tasks, which brings a big challenge for PLMs to transfer the knowledge from the source task to the target task.}, such as the natural language inference (NLI) datasets MNLI~\cite{Williams2018A} and SNLI~\cite{Bowman2015A}.
To accommodate both learning settings,~\emph{\model} employs a unified learning paradigm, with the high-level architecture shown in Figure~\ref{fig:overview}.
In \emph{\model}, we first employ a~\emph{Multi-task Meta-knowledge Acquisition} (MMA) procedure to learn the transferable representations of prompt encoders across multiple tasks. 
For learning across similar tasks, multiple task-specific prompt encoders are used to capture the private knowledge of each task, while a universal prompt encoder is exploited to learn the shared semantics across tasks. 
For learning across distant tasks, we additionally introduce task type descriptions with pseudo templates (which can be viewed as the task-wise knowledge) and then calculate the intra-type and inter-type prompt embeddings to make PLMs better mine implicit relations among multiple distant tasks.
To reduce {over-fitting} and make the underlying PLM {more task-agnostic} and {less unbiased} towards any specific tasks, we further propose two~\emph{de-biasing} techniques, namely~\emph{prototype-based de-biasing} and~\emph{entropy-based de-biasing}. 
The learned model can be viewed as a~\emph{meta-learner} for a group of tasks.

After MMA,~\emph{\model} takes the~\emph{Task-aware Model Specification} (TMS) step, which can be further divided into two cases. 
i) When the model is adapted to existing tasks during MMA, 
we directly continually tune the corresponding prompt encoder on its own dataset.
ii) When it is required to fit a previously unseen task, a~\emph{model generalization} strategy is employed, specifically considering the universal prompting knowledge in the model. This is often the case where re-training of the~\emph{meta-leaner} across all the tasks is infeasible due to data privacy or computation efficiency issues.

For evaluation, we test the~\emph{\model} framework on three types of few-shot NLP tasks (including seven public datasets in total): i) sentiment analysis, ii) NLI, and iii) paraphrasing. Experimental results show that~\emph{\model} consistently outperforms both single-task and cross-task strong baselines. We further show that i) the meta-learner trained by~\emph{\model} is effective to generalize to unseen tasks, and ii)~\emph{\model} also outperforms popular fine-tuning algorithms when learning with full training sets.

In summary, the major contributions of this work are:
\begin{itemize}
\item We introduce the novel \emph{\model} framework for text classification to learn transferable knowledge across either similar tasks and distant tasks.

\item A prompt-based meta-learner training algorithm with two de-biasing techniques is presented to improve the model adaptation and generalization.

\item Experiments on multiple types of text classification tasks show that the proposed~\emph{\model} framework consistently outperforms strong baselines in both few-shot learning setting and full-data scenario.
\end{itemize}

The rest of this paper is organized as follows. Section~\ref{sec:re} summarizes the related work. 
Details of~\emph{\model} and experimental results are presented in Section~\ref{sec:model} and Section~\ref{sec:exp}. 
Finally, we 
draw a conclusion in Section~\ref{sec:con}.

\section{Related Work}
\label{sec:re}

In this section, we summarize the related work on PLMs, prompt-based learning for text classification, transfer learning and meta-learning.

\subsection{Pre-trained Language Models (PLMs)}
With large-scale pre-training, PLMs have achieved significant improvements over various NLP tasks through the task-specific fine-tuning.
For example, BERT~\cite{Devlin2019BERT} captures contextual semantic representations by multi-layer transformer encoders~\cite{Vaswani2017Attention}, and is pre-trained with two self-supervised learning objectives, i.e., Masked Language Modeling (MLM) and Next Sentence Prediction (NSP). 
Recently, a series of transformer encoder-based methods are proposed to improve the performance of the vanilla BERT model, including RoBERTa~\cite{Liu2019RoBERTa}, ALBERT~\cite{Lan2020ALBERT}, XLNet~\cite{Yang2019XLNet}, DeBERTa~\cite{He2021Deberta}, etc. 
Additionally, structured knowledge graphs can also be used to enhance PLMs~\cite{Xiong2020Pretrained, Liu2020K-BERT, Zhang2019ERNIE, Wang2021KEPLER}. For example, WKLM~\cite{Xiong2020Pretrained} employs entity-level masking and replacement strategies to enhance the representations of tokens by learning from the explicit knowledge in knowledge graphs. KEPLER~\cite{Wang2021KEPLER} integrate the pre-trained knowledge base embeddings (KBE) into PLMs through well-designed attentive fusion modules to improve the contextual representations. 
Rather than using transformer encoders only, the encoder-decoder architecture and the decoder-only architecture have been employed to make PLMs better adapt to both natural language understanding and generation tasks, such as BART~\cite{Lewis2020BART}, T5~\cite{Raffel2020Exploring}, etc. 
As the neural architecture design of PLMs is not our focus, we do not further elaborate.

\subsection{Prompt-based Learning for Text Classification}
For text classification tasks, directly fine-tuning the classification head over PLMs may perform poorly in low-resource settings~\cite{liu2021pre}. Recently, the huge GPT-3 model~\cite{Brown2020Language} has been proposed to enable in-context learning by handcrafted templates and demonstrations. To apply in-context learning to BERT-style models,
Schick and Sch{\"{u}}tze~\cite{Schick2021Exploiting} introduce handcrafted Pattern-Verbalizer-Pair (PVP) to unify all text classification into cloze-style formats.
To facilitate the automatic PVP generation, Gao et al.~\cite{Gao2021Making} present LM-BFF to generate discrete templates and label word mappings~\cite{Raffel2020Exploring}. Other works~\cite{Shin2020AutoPrompt, han2021ptr} mine prompts from the training corpus based on heuristic rules/semantic relations. However, these methods are time-consuming for mining optimized prompts for target tasks. A series of methods are proposed to learn continuous/soft prompt embeddings, such as P-tuning~\cite{Liu2021GPT}, P-tuning-V2~\cite{liu2021ptuningv2}, OptiPrompt~\cite{Zhong2021Factual}, Prefix-tuning~\cite{Li2021Prefix}. Zhao et al.~\cite{Zhao2021Discrete} and Gu et al.~\cite{Gu2021PPT} focus on hybrid training with both discrete and continuous prompts. He et al.~\cite{Hu2021Knowledgeable} considers the automatic expansion of label words and presents Knowledgeable Prompt-tuning (KPT) to utilize knowledge for the construction of verbalizers. 
In addition, 
a few works~\cite{Li2021SentiPrompt, Chen20210KnowPrompt} focus on prompts for specific tasks, such as sentiment analysis~\cite{Li2021SentiPrompt}, relation extraction~\cite{Chen20210KnowPrompt}.
In this paper, we further leverage data from non-target tasks to make prompt-tuned PLMs have better capacities for adapting to unseen tasks.


\subsection{Transfer Learning and Meta-learning}
Transfer learning aims to transfer knowledge or resources from source domains to target domains~\cite{Pan2010A, Lu2015Transfer, Wang2019A}. 
For deep neural networks, it is common practice to learn similar tasks by multi-task learning~\cite{Liu2019Multi}. With the popularity of PLMs, fine-tuning has become the standard practice by learning from PLMs for similar tasks~\cite{Sun2019How,Arase2021Transfer}.
In contrast, meta-learning aims to learn models that can quickly adapt to different tasks with little training data available~\cite{Huang2022Meta}, typically formulated as an~\emph{N-way K-shot} problem. Meta-learning algorithms have been applied in few-shot NLP tasks, such as text classification~\cite{Geng2020Dynamic}, information extraction~\cite{Gao2019FewRel} and question answering~\cite{Hua2020Few}, etc.
We notice that our proposed~\emph{\model} framework is not a typical~\emph{N-way K-shot} paradigm. Similar to~\cite{Wang2020Meta,Wang2021MeLL}, ~\emph{\model} can be viewed as a combination of transfer learning and meta-learning, which learns transferable knowledge from either similar or distant tasks to improve the performance of few-shot text classification.

\begin{figure*}
\centering
\includegraphics[width=\textwidth]{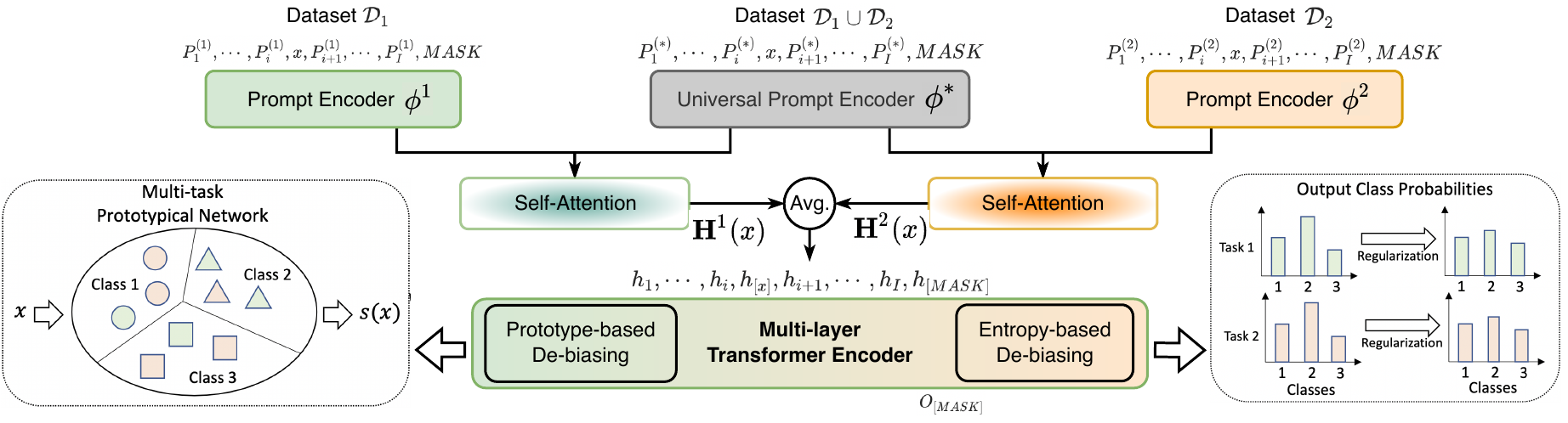}
\caption{The model architecture of the meta-learner training process during multi-task learning across \textbf{similar} tasks. For simplicity in this figure, we assume there are two tasks and three classes for few-shot text classification.
The multi-domain few-shot datasets are passed through task-specific and universal prompt encoders. After that, a multi-layer transformer encoder (e.g., BERT, RoBERTa, etc.) is employed to capture the transferable knowledge across tasks, with two de-biasing techniques proposed.
(Best viewed in color.) 
}
\label{fig:meta}
\end{figure*}

\section{The \emph{\model} Framework}
\label{sec:model}


\subsection{Overview}

\subsubsection{Task Description}

We introduce a summary the cross-task settings. 
Let $\mathcal{T}_1,\cdots,\mathcal{T}_M$ be $M$ source tasks with very few labeled instances. The $m$-th task can be formulated as: $\mathcal{T}_m:x\rightarrow y$, where $x$ and $y\in\mathcal{Y}_{m}$ represent the input text~\footnote{Note that the input $x$ can be either a single sentence or a sentence pair (which has the same setting as that of BERT~\cite{Devlin2019BERT}). For simplicity, we uniformly denote the input as $x$ throughout this paper.} and the classification label, respectively. $\mathcal{Y}_m$ is the classification label set of the task $\mathcal{T}_m$. 
In our setting, we assume that there are $K$ training samples associated with each class $y\in\mathcal{Y}_{m}$. Hence, for each task $\mathcal{T}_m$, we have a training set $\mathcal{D}_m$ with $\vert\mathcal{Y}_{m}\vert\times K$ training samples, where $\vert\mathcal{Y}_{m}\vert$ denotes the class number of the task $\mathcal{T}_m$. 
The total number of training instances of $M$ tasks is $K\cdot\sum_{m}\vert \mathcal{Y}_m\vert$.

\subsubsection{Model Overview}

In~\emph{\model}, we first train a~\emph{meta-learner}~$\mathcal{F}_{meta}$ over $M$ few-shot training sets $\mathcal{D}_1,\cdots,\mathcal{D}_M$ with parameters initialized from any PLMs. We denote this training process as the stage of \emph{Multi-task Meta-knowledge Acquisition} (MMA). As mentioned above, we consider two scenarios. 1) Training across similar tasks. In this case, the selected $M$ training tasks share the same label space. Formally, $\forall\mathcal{T}_i, \mathcal{T}_j, 1\leq i\leq j\leq M$, we have $\mathcal{Y}_i=\mathcal{Y}_j$.
2) Training across distant tasks. The selected $M$ training sets can have different label spaces. This means, there exist at least two tasks $\mathcal{T}_i$ and $\mathcal{T}_j$ in $\mathcal{T}_1,\cdots,\mathcal{T}_M$ such that $\mathcal{Y}_i\neq\mathcal{Y}_j$. We denote $R$ as the number of task types, all the tasks with the same $r$-th type~\footnote{We suppose that all the tasks with the same label space and similar data distribution belong to the same task type, which can be viewed as the similar tasks; otherwise, they are distant tasks.} can be grouped into $\mathcal{G}_r$, where contains $M_r$ tasks. Thus, $\sum_{r=1}^RM_r=M$.
For example, suppose that the source data derived from one sentiment analysis task (e.g., SST-2) and two NLI tasks (e.g., MNLI and SNLI), we have $R=2$, $M=3$, $M_1=1$ and $M_2=2$.

After training the meta-learner, we propose~\emph{Task-aware Model Specification} (TMS) to further tune the meta-learner $\mathcal{F}_{meta}$ on the given target task. 
We consider two real-world scenarios, i.e., model adaptation and model generalization:
\begin{enumerate}
    \item \textbf{Model adaptation}. $\mathcal{F}_{meta}$ can be adapted to each seen target task $\mathcal{T}_m\in\mathcal{T}_1, \cdots, \mathcal{T}_M$ based on its own training set $\mathcal{D}_m$.
    \item \textbf{Model generalization}. 
    As $\mathcal{F}_{meta}$ is designed to digest the~\emph{transferable knowledge} across tasks, rather than simple multi-task learning, $\mathcal{F}_{meta}$ can also be generalized to previously unseen tasks. We denote this unseen task as $\mathcal{\tilde T}$, and $\mathcal{\tilde T}\notin\mathcal{T}_1, \cdots, \mathcal{T}_M$. Due to the data privacy or computation efficiency issues, when the few-shot training set $\mathcal{\tilde D}$ of the $\mathcal{\tilde T}$ is not available during the training process of $\mathcal{F}_{meta}$, we explore how~\emph{\model} can be used to generate an accurate model $\mathcal{\tilde F}$ based on $\mathcal{F}_{meta}$.
In this case, $\mathcal{F}_{meta}$ does not have any knowledge of the new task $\mathcal{\tilde T}$ when in the cross-task training stage.
\end{enumerate}

In the following, 
we elaborate on the two major stages of the~\emph{\model} framework, i.e.,~\emph{Multi-task Meta-knowledge Acquisition} (MMA) and~\emph{Task-aware Model Specification} (TMS).
After that, we also discuss how to apply~\emph{\model} to standard fine-tuning scenarios where we have relatively large training sets.

\subsection{Multi-task Meta-knowledge Acquisition (MMA)}

During the MMA stage, the PLM can learn meta-knowledge across multiple source tasks with a cross-task learning process~\cite{Wang2020Meta, Wang2022Towards}. As mentioned above, if all the source tasks share the same label space, we consider this scenario as \emph{learn across similar task}. Likewise, if there existing two source tasks have different label spaces, we consider this situation as \emph{learn across distant task}.

\noindent\textbf{Prompt-based Learning across Similar Tasks.}
We consider the first situation where~\emph{\model} aims to transfer prompt learning across similar tasks.
For clarity, the general architecture of the meta-learner for similar-task learning is illustrated in Figure~\ref{fig:meta}.
For prompt encoding, we augment an input sentence $x$ with some pseudo template tokens. The prompt template $t(x)$ is shown as follows:
\begin{equation*}
t(x)=P_1,\cdots,P_i,x,P_{i+1},\cdots,P_I,MASK,
\end{equation*}
where $P_i$ denotes the prompt pseudo token and can be viewed as a continuous vector (as proposed in~\cite{Liu2021GPT}), $I$ is the total number of pseudo tokens, and $MASK$ is a special masked language token as the placeholder for model output. 
Then, we employ a prompt encoder $\text{PE}(\cdot)$ to capture the dependency of the pseudo tokens in the input sequence, which contains one bidirectional LSTM network with multi-layer perceptrons (MLP).
Formally, we have:
\begin{align*}
\text{PE}_{\phi}(t(x))= \text{MLP}_{\phi}\big (\text{BiLSTM}_{\phi}(t(x))\big ),
\end{align*}
where $\phi$ denotes trainable parameters of the prompt encoder.


As the~\emph{\model} framework is placed in the multi-task setting, for each task $\mathcal{T}_m$, we have a task-specific prompt template $t^{(m)}(x)$ as follows:
\begin{equation*}
t^{(m)}(x)=P_1^{(m)},\cdots,P_i^{(m)},x,P_{i+1}^{(m)},\cdots,P_I^{(m)},MASK.
\end{equation*}
Besides, we also define a universal prompt template $t^{(*)}(x)$ for all tasks:
\begin{equation*}
t^{(*)}(x)=P_1^{(*)},\cdots,P_i^{(*)},x,P_{i+1}^{(*)},\cdots,P_I^{(*)},MASK.
\end{equation*}

For an instance $(x,y)\in\mathcal{D}_m$, the prompt embedding $\mathbf{H}^{m}(x)$ can be computed as follows:
\begin{align*}
\mathbf{H}^{m}(x)= \text{SelfAtt} \big(
\text{PE}_{\phi^{m}}(t^{(m)}(x)), \text{PE}_{\phi^{*}}(t^{(*)}(x)) \big),
\end{align*}
where $\phi^{m}$ and $\phi^{*}$ denote the parameters of the task-specific prompt encoder and the universal prompt encoder, respectively. $\text{SelfAtt}(\cdot, \cdot)$ is the self-attention pooling layer~\cite{Vaswani2017Attention} which aims at obtaining the fusion prompt embedding $\mathbf{H}^{m}(x)$ from both task-specific and universal prompt encoders. 
Finally, we perform average pooling over multiple prompt embeddings of all tasks, the final prompt embedding is a sequence as the input of the PLM, i.e.,
\begin{equation*}
\mathbf{H} = h_1,\cdots,h_i,h_{[x]},h_{i+1},\cdots,h_{I},h_{[MASK]}
\end{equation*}
where $h_{[x]}$ is the sequence embedding of input $x$, and $h_{[MASK]}$ is the masked output token embedding. 
As the parameters of the prompt encoders are fully differentiable, during backpropagation, they can effectively capture task-specific and universal knowledge.

\begin{figure*}
\centering
\includegraphics[width=\textwidth]{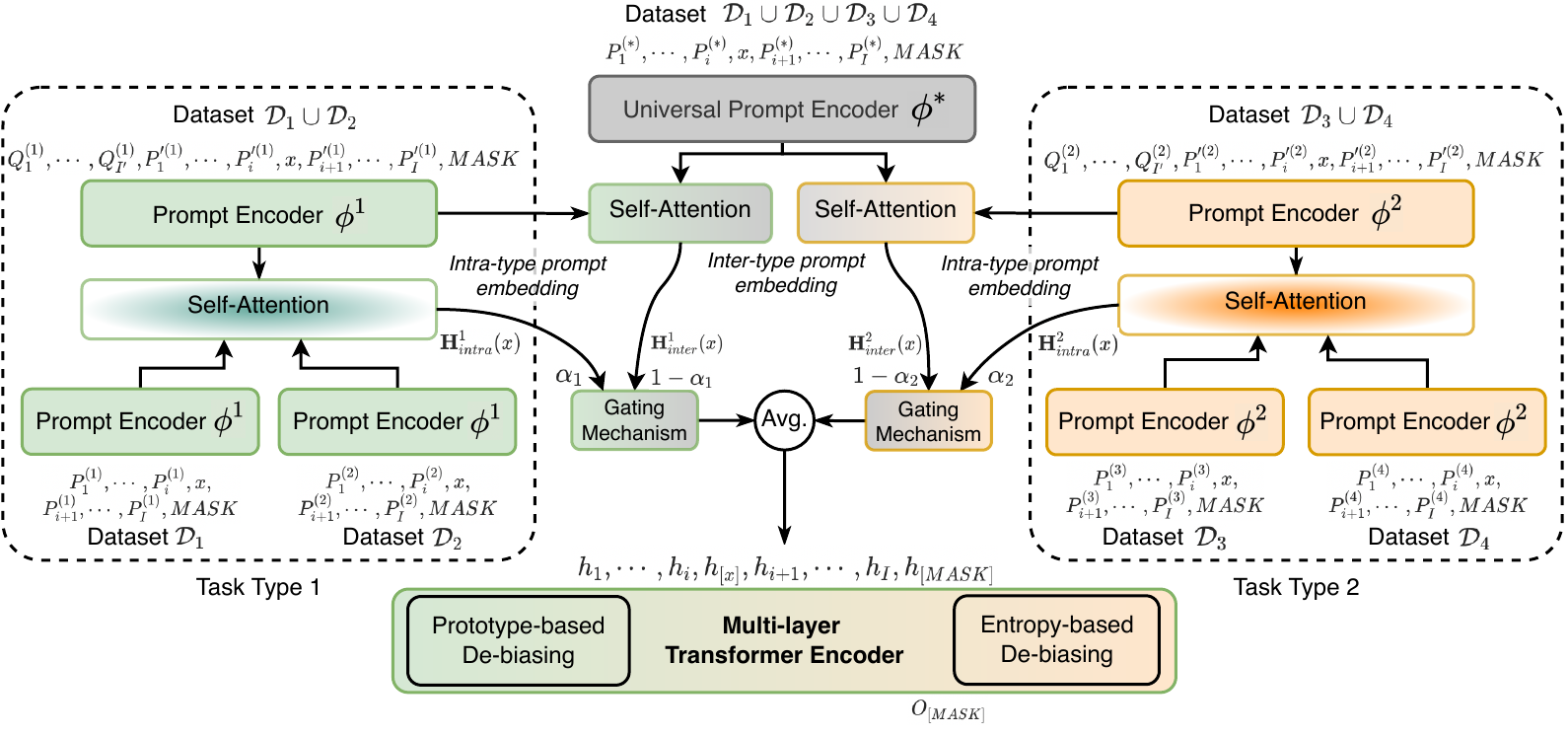}
\caption{The model architecture of the meta-learner training process during multi-task learning across \textbf{distant} tasks. 
Suppose that we have two task types and each containing two tasks. 
For each type, we can obtain intra-type prompt embeddings. All input data with the same task type are injected with the corresponding type description. We then obtain the inter-type prompt embeddings based on the task-specific and universal prompt encoders.
Two de-biasing techniques are also used for the model improvement, we omit the details of them in this figure.
(Best viewed in color.) 
}
\label{fig:meta_distant}
\end{figure*}

\noindent\textbf{Prompt-based Learning across Distant Tasks.}
By intuition, PLMs trained over tasks with similar label spaces can learn the transferable knowledge across these tasks. However, it is difficult for the previous prompt encoding approach to directly apply to various tasks with different label spaces. In addition, the huge difference in the intrinsic semantic distributions among these distant tasks makes PLMs difficult to learn the implicit mutual relations. Hence, we extend our prompt encoding method to distant NLP tasks. The architecture is shown in Figure~\ref{fig:meta_distant}.

To make PLMs understand the task semantics more clearly, we design concise text descriptions for each task type and concatenate them with the corresponding prompts.
Specifically, for the $r$-th task type, we have a group of tasks $\mathcal{G}_r$. Given one task $\mathcal{T}_{m}\in\mathcal{G}_r$, for each instance $(x, y)\in\mathcal{D}_m$, the task-specific and universal prompt templates are respectively denoted as $t^{(m)}(x)$ and $t^{(*)}(x)$, which are the same as described previously.
We further introduce a new prompt with the task type description denoted as $t'^{(r)}(x)$, where instances with the same task type share the same template, represented as:
\begin{equation*}
Q_1^{(r)},\cdots, Q_{I'}^{(r)}, P_1'^{(r)},\cdots,P_i'^{(r)},x,P_{i+1}'^{(r)},\cdots,P_I'^{(r)},MASK.
\end{equation*}
where $Q_i^{(r)}$ refers to the text tokens of the description, $I'$ denotes the length of the description. $P_i'^{(r)}$ is the pseudo token. 

Different from the previous approach, we consider two aspects of prompt encoding, include the intra-type prompt embedding $\mathbf{H}^{r}_{intra}(x)$ and the inter-type prompt embedding $\mathbf{H}^{r}_{inter}(x)$, defined as follows:
\begin{align*}
\mathbf{H}^{r}_{intra}(x)= \frac{1}{M_r}\sum_{m=1}^{M_r}\text{SelfAtt} \big(
\text{PE}_{\phi^{r}}(t^{(m)}(x)), \text{PE}_{\phi^{r}}(t'^{(r)}(x)) \big),
\end{align*}
\begin{align*}
\mathbf{H}^{r}_{inter}(x)= \text{SelfAtt} \big(
\text{PE}_{\phi^{r}}(t'^{(r)}(x)), \text{PE}_{\phi^{*}}(t^{(*)}(x)) \big),
\end{align*}
where $\phi^{r}$ and $\phi^{*}$ denote the parameters of task type-specific and universal prompt encoders, respectively.
In this manner, for one batch of examples, the parameters of $\text{PE}_{\phi^{r}}(\cdot)$ can be optimized by the all task data that share the same type in $\mathcal{G}_{r}$. At the same time, the universal prompt encoder $\text{PE}_{\phi^{*}}(\cdot)$ aims to learn the universal knowledge from all types of examples. 
Finally, we directly combine the two prompt embeddings by a simple gating mechanism:
\begin{align*}
\mathbf{H}^{r}(x)= \alpha_r \mathbf{H}^{r}_{intra}(x) + (1 - \alpha_r) \mathbf{H}^{r}_{inter}(x),
\end{align*}
where $\alpha_r$ is the trainable balancing coefficient, with $0<\alpha_r<1$. We perform average pooling over prompt embeddings of all types and obtain the final prompt embeddings $\mathbf{H}$ as the input of the PLM. 

Finally, we can obtain the output representations $O_{MASK}$ at the position of $MASK$ token, which can be viewed as the sentence embedding $\mathcal{E}(x)$.

\subsection{The Training Process of the Meta-learner}

During model pre-training, we regard the prompt embedding $\mathbf{H}$ as the model input and train the meta-learner via multi-task learning over the MLM head.
As large PLMs can easily suffer from over-fitting during few-shot learning~\cite{Gao2021Making}, in cross-task scenarios, the meta-learner would unfortunately memorize the ~\emph{non-transferable knowledge} from non-target tasks. 
To alleviate this problem, we propose two de-biasing techniques to obtain a more~\emph{unbiased} meta-learner encoded with~\emph{transferable knowledge}, namely i)~\emph{prototype-based de-biasing} and ii)
~\emph{entropy-based de-biasing}. 


\noindent{\bf Prototype-based De-biasing.}
This technique aims to give more importance to~\emph{prototypical instances across tasks} during the training process of the meta-learner. Here, we extend~\cite{Snell2017Prototypical} to construct a lite~\emph{Multi-task Prototypical Network}.
In the network, the task-specific class centroid embedding $\mathbf{c}_m(y)$ ($y\in\mathcal{Y}_m$) w.r.t. the task $\mathcal{T}_m$ is computed and stored as:
\begin{equation*}
\mathbf{c}_m(y)=\frac{1}{\vert\mathcal{D}_{m,y}\vert}\sum_{(x,y)\in\mathcal{D}_{m,y}}\mathcal{E}(x) 
\end{equation*}
where $\mathcal{D}_{m,y}$ is the subset of $\mathcal{D}_{m}$ such that each instance in $\mathcal{D}_{m,y}$ has the label $y$, and $\mathcal{E}(x)$ is the representation of $x$ generated by the meta-learner described previously.
For each instance $(x,y)\in\mathcal{D}_m$, we pass the text $x$ through the network to generate the~\emph{cross-task prototype score}, denoted as $s(x)$. Formally, we have:
\begin{equation*}
\begin{split}
& s(x)=\zeta\cdot\frac{\text{sim}(\mathcal{E}(x),\mathbf{c}_m(y))}{\sum_{\tilde y\in\mathcal{Y}_m}\text{sim}(\mathcal{E}(x),\mathbf{c}_m(\tilde y))}\\
& +\frac{1-\zeta}{M-1}\sum_{\tilde m=1 (m\neq\tilde m)}^{M}\frac{\text{sim}(\mathcal{E}(x),\mathbf{c}_{\tilde m}(y))}{\sum_{\tilde y\in\mathcal{Y}_{\tilde{m}}}\text{sim}(\mathcal{E}(x),\mathbf{c}_{\tilde m}(\tilde y))}.
\end{split}
\end{equation*}
where $0<\zeta<1$ is a pre-defined balancing factor, and $\text{sim}(\cdot,\cdot)$ is the similarity function between two embeddings.
We can see that an instance receives a higher score if it is semantically related to the centroids from both the task $\mathcal{T}_m$ itself and other tasks, hence is~\emph{more transferable across tasks}. 

By treating $s(x)$ as the optimization weight, the overall loss function $\mathcal{L}(\Theta)$ of  $\mathcal{F}_{meta}$ can be given by:
\begin{equation*}
\mathcal{L}(\Theta)=\sum_{m=1}^{M}\sum_{(x,y)\in\mathcal{D}_m}s(x) f(x,y;\Theta)+\lambda_1\Vert\Theta\Vert,
\end{equation*}
where $\Theta$ is the collection of all model parameters, $f(x,y;\Theta)$ is the~\emph{sample-wise cross-entropy loss}, and $\lambda_1$ is the regularization hyper-parameter.

\noindent{\bf Entropy-based De-biasing.}
One potential risk of applying the~\emph{prototype-based de-biasing} technique only is obtaining a~\emph{non task-agnostic} meta-learner. Consider three tasks $\mathcal{T}_1$, $\mathcal{T}_2$ and $\mathcal{T}_3$. If $\mathcal{T}_1$ and $\mathcal{T}_2$ are highly similar, and $\mathcal{T}_3$ is more dissimilar. Instances in $\mathcal{D}_1$ and $\mathcal{D}_2$ would naturally receive high prototype scores, making the meta-learner~\emph{biased} towards $\mathcal{T}_1$ and $\mathcal{T}_2$, and pays little attention to $\mathcal{T}_3$. Hence, when the meta-learner is required to fit $\mathcal{T}_3$, it may have poor parameter initialization settings. To make it~\emph{more task-agnostic}, inspired by~\cite{Jamal2019Task}, we consider the model prediction entropy $\mathcal{H}(\mathcal{D}_m)$ over $\mathcal{D}_m$:
\begin{equation*}
\mathcal{H}(\mathcal{D}_m)=-\frac{1}{\vert\mathcal{D}_m\vert}\sum_{(x,y)\in \mathcal{D}_m}\sum_{\hat{y}\in\mathcal{Y}_{*}}\hat{y}(x)\log\hat{y}(x),
\end{equation*}
where $\hat{y}(x)$ is the predicted probability of $x$ being assigned to the class $\hat{y}\in\mathcal{Y}$. If we learn across similar tasks, we have $\mathcal{Y}_{*}=\mathcal{Y}$; otherwise, $\mathcal{Y}_{*}=\mathcal{Y}_m$. We can see that when $\mathcal{H}(\mathcal{D}_m)$ is used as a part of the model regularizers, the meta-learner will be~\emph{less over-trained} on any specific tasks.

By plugging the term $\mathcal{H}(\mathcal{D}_m)$ into the loss function $\mathcal{L}(\Theta)$, we have the new loss function $\mathcal{L}^{'}(\Theta)$:
\begin{equation*}
\begin{split}
\mathcal{L}^{'}(\Theta)= & \sum_{m=1}^{M}\sum_{(x,y)\in\mathcal{D}_m}(s(x) f(x,y;\Theta)\\
& -\frac{\lambda_2}{\vert\mathcal{D}_m\vert}\sum_{\hat{y}\in\mathcal{Y}_{*}}\hat{y}(x)\log\hat{y}(x))+\lambda_1\Vert\Theta\Vert,
\end{split}
\end{equation*}
where $\lambda_2$ is the regularization hyper-parameter.

\begin{algorithm}[t]
\caption{Meta-learner Training Algorithm}
\label{alg:meta}
\begin{algorithmic}[1]
\FOR {each instance $(x,y)\in\bigcup_{m=1}^{M}\mathcal{D}_m$}
\STATE Uniformly set $s(x)=1$;
\ENDFOR
\WHILE {number of training epochs does not reach a limit}
\WHILE {current training epoch is not finished}
\STATE Sample a batch $\mathcal{B}=\{(x,y)\}$ from $\bigcup_{m=1}^{M}\mathcal{D}_m$;
\STATE Use $\mathcal{B}$ to update $\mathcal{F}_{meta}$ by minimizing $\mathcal{L}^{'}(\Theta)$;
\ENDWHILE
\FOR {each instance $(x,y)\in\bigcup_{m=1}^{M}\mathcal{D}_m$}
\STATE Compute $s(x)$ based on the updated model;
\ENDFOR
\ENDWHILE
\STATE \textbf{return} the meta-learner $\mathcal{F}_{meta}$ (i.e., parameters of the PLM and multiple prompt encoders).
\end{algorithmic}
\end{algorithm}

\noindent{\bf Optimization Procedure.}
Despite its simple formula, minimizing $\mathcal{L}^{'}(\Theta)$ is a non-trivial problem. This is because when we calculate $s(x)$, we must obtain model parameters of the PLM beforehand, which is not available before the training process. On the other hand, the optimization of $\mathcal{L}^{'}(\Theta)$ requires the values of $s(x)$ for all training samples, which poses the~\emph{``chicken-and-egg''} problem.

We employ a~\emph{dual optimization} process to solve the problem of $\mathcal{L}^{'}(\Theta)$. In the initial stage, all $s(x)$s are uniformly initialized. Next, we fix $s(x)$ scores as constants to minimize $f(x,y;\Theta)$ in $\mathcal{L}^{'}(\Theta)$. An inference procedure on the PLM can be applied to obtain all $s(x)$ scores. This process iterates for a certain number of epochs. Readers can also refer to Algorithm~\ref{alg:meta} for an algorithmic overview.

\subsection{Task-aware Model Specification (TMS)}

After MMA, the meta-learner can be adapted to specific tasks with ease. For a target task $\mathcal{T}_m$ that has already ``seen'' by the meta-learner, we fine-tune the corresponding task-specific prompt encoder and the PLM by minimizing the loss function $\mathcal{L}^{(m)}(\Theta)$:
\begin{equation*}
\mathcal{L}^{(m)}(\Theta)=\sum_{(x,y)\in\mathcal{D}_m} f(x,y;\Theta)+\lambda_1\Vert\Theta\Vert,
~\label{eqn:loss}
\end{equation*}

For a previously unseen target task $\mathcal{\tilde T}$, the~\emph{model generalization} strategy is employed.
Here, we first generate a task-specific template (consisting of both type description tokens and pseudo tokens) for this target task and then use the universal prompt encoder to initialize its prompt encoder. The entire model is trained over the dataset $\mathcal{\tilde D}$, with the loss function $\mathcal{\tilde L}(\Theta)$ defined as follows:
\begin{equation*}
\mathcal{\tilde L}(\Theta)=\sum_{(x,y)\in\mathcal{\tilde D}} f(x,y;\Theta)+\lambda_1\Vert\Theta\Vert,
~\label{eqn:loss2}
\end{equation*}
As the meta-learner is highly generalized, it can provide good initialization for the few-shot learning task $\mathcal{\tilde T}$.

\subsection{Learning with Full Training Sets}
\emph{\model} can also be applied for standard fine-tuning when we have relatively large training sets with few modifications.
During MMA, we notice that when it is not an~\emph{N-way K-shot} problem, the sizes of $\mathcal{D}_1,\cdots,\mathcal{D}_M$ can be significantly different. Optimizing $\mathcal{L}^{'}(\Theta)$ directly on these datasets would make the meta-learner~\emph{biased towards large datasets}. To remedy this dilemma, when we sample a batch from $\mathcal{D}_1,\cdots,\mathcal{D}_M$, instead of randomly selection, we employ stratified sampling where training instances are selected with the probability proportional to the dataset distribution $\Pr(\mathcal{D}_m)$:
\begin{equation*}
\Pr(\mathcal{D}_m)=\frac{\log\vert\mathcal{D}_m\vert+\gamma}{\sum_{\tilde m=1}^{M}\log\vert\mathcal{D}_{\tilde m}\vert+\gamma},
\end{equation*}
where $\gamma>0$ is a smoothing factor. This results in the~\emph{over-sampling} of small datasets and the~\emph{under-sampling} of large datasets, which helps to improve the generalization capability of the model.

\section{Experiments}
\label{sec:exp}

In this section, we conduct extensive experiments to evaluate the~\emph{\model} framework and compare it against strong baselines.


\begin{table*}
\centering
\caption{The specific information of each dataset. \texttt{[<s1>]} and \texttt{[<s2>]} are the original sentences, \texttt{[MASK]} is the maked token, and \texttt{[p$_i$]} is the pseudo token.} 
\begin{small} 
\begin{tabular}{c | c | c | c | l | l}  
\midrule
\bf Task Type (Group) & \bf Task & \bf \#Train & \bf \#Test & \bf Template Format & \bf Task Type Description Example \\
\midrule
\multirow{3}{*}{Sentiment Analysis} & SST-2 & 6,920 & 872 & \texttt{[<s1>]}. \texttt{[p1]} $\cdots$ \texttt{[MASK]}. & \multirow{3}{*}{A sentiment analysis task.} \\
& MR & 8,662 & 2,000 & \texttt{[<s1>]}. \texttt{[p1]} $\cdots$ \texttt{[MASK]}. &  \\
& CR & 1,775 & 2,000 & \texttt{[<s1>]}. \texttt{[p1]} $\cdots$ \texttt{[MASK]}. & \\
\midrule

\multirow{2}{*}{NLI} & MNLI & 392,702 & 9,815 & \texttt{[<s1>]}. \texttt{[p1]} $\cdots$ \texttt{[MASK]}. \texttt{[<s2>]}. & \multirow{2}{*}{A natural language inference.} \\
& SNLI & 549,367 & 9,842 & \texttt{[<s1>]}. \texttt{[p1]} $\cdots$ \texttt{[MASK]}. \texttt{[<s2>]}. &  \\
\midrule

\multirow{2}{*}{paraphrasing} & MRPC & 3,668 & 408 & \texttt{[<s1>]}. \texttt{[p1]} $\cdots$ \texttt{[MASK]}. \texttt{[<s2>]}. & \multirow{2}{*}{A paraphrasing task.} \\
& QQP & 363,846 & 40,431 & \texttt{[<s1>]}. \texttt{[p1]} $\cdots$ \texttt{[MASK]}. \texttt{[<s2>]}. &  \\
\midrule
\end{tabular} 
\end{small}
\label{tab:dataset}
\end{table*}

\subsection{Datasets and Experimental Settings}
Following~\cite{wang2021transprompt,Gao2021Making,Liu2021GPT},
we select seven public datasets to evaluate~\emph{\model}, divided into three groups of NLP tasks: sentiment analysis (SST-2~\cite{Socher2013Socher}, MR~\cite{Hu2004Mining} and CR~\cite{Pang2005Seeing}), NLI (MNLI~\cite{Williams2018A} and SNLI~\cite{Bowman2015A}) and paraphrasing (MRPC~\cite{Dolan2005Dolan} and QQP~\footnote{\url{https://www.quora.com/q/quoradata/}.}).
Note that we do not use other benchmark datasets (such as CoLA) because datasets on similar tasks to CoLA are not available to us. Hence, it is impossible to train the meta-learner in~\emph{\model}.
We provide the template format and task type description for each task. For the verbalizer, we follow~\cite{wang2021transprompt, Gao2021Making} to use the top-3 label mapping and report the average results in experiments.
The detailed information of these datasets is reported in Table~\ref{tab:dataset}. The training/development/testing splits are the same as~\cite{wang2021transprompt}.

For few-shot learning, the evaluation protocols are the same as~\cite{Gao2021Making, wang2021transprompt}. The underlying PLM is the RoBERTa large model (with 335M parameters)~\cite{Liu2019RoBERTa} and we set $K=16$. We measure the
average performance in terms of accuracy across 5 different randomly
sampled training and development splits. Refer to~\cite{Gao2021Making} for more experimental settings.
We employ the following methods as single-task baselines:
\begin{itemize}
    \item Standard BERT-style~\cite{Devlin2019BERT} fine-tuning\footnote{\url{https://github.com/huggingface/transformers}.};
    \item The LM-BFF prompting model~\cite{Gao2021Making} (with both manually-compiled and automatically-mined prompts)\footnote{\url{https://github.com/princeton-nlp/LM-BFF}.};
    \item P-tuning~\cite{Liu2021GPT}, which produces state-of-the-art performance for few-shot learning based on continuous prompt embeddings\footnote{\url{https://github.com/THUDM/P-tuning}.}.
\end{itemize}

Because we focus on learning transferable prompting knowledge across tasks, we also use the multi-task version baselines of BERT fine-tuning~\cite{Sun2019How}, LM-BFF~\cite{Gao2021Making} and P-tuning~\cite{Liu2021GPT} for learning across both similar and distant tasks. Specifically, we employ separate prompts (either discrete prompts or continuous prompt embeddings) for different tasks in the multi-task versions of LM-BFF and P-tuning.
In addition, Meta Fine-tuning (MFT)~\cite{Wang2020Meta} is employed as a baseline for similar tasks, while the Meta Distant Transfer Learning framework (Meta-DTL)~\cite{Wang2021Meta} is used as a baseline for learning across distant tasks.
During MMA from similar tasks, we constrain that the knowledge is transferred within the same group of NLP tasks. For example, the meta-learner for sentiment analysis is jointly trained over the few-shot training sets of SST-2, MR, and CR. 
When training across distant tasks, apart from the target task (e.g., SST-2), we also select other types of datasets (e.g., NLI and paraphrasing) for multi-task learning. After multi-task training, only training data of the target task is used to perform task-specific prompt-tuning.

\begin{table*}
\centering
\caption{\label{tab:few-shot}
The few-shot testing results of~\emph{\model} and baselines in terms of accuracy (\%).
``man'', ``auto'' and ``mtl'' refer to manually-compiled prompts, automatically-mined prompts, and multi-task learning, respectively.
$^{*}$ refers to the multi-task variants of the original approaches.
Hereinafter the same.
}
\begin{small}
\begin{tabular}{l | ccc | cc | cc | c }
\midrule
\multirow{2}{*}{\textbf{Method}} & \multicolumn{3}{c|}{\bf Task: Sentiment Analysis} & \multicolumn{2}{c|}{\bf Task: NLI} & \multicolumn{2}{c|}{\bf Task: paraphrasing} & \multirow{2}{*}{\textbf{Average}}\\
\cmidrule{2-8}
 & \textbf{SST-2} & \textbf{MR} & \textbf{CR} & \textbf{MNLI} & \textbf{SNLI} & \textbf{MRPC} & \textbf{QQP}\\
\midrule
\multicolumn{8}{l}{\em Single-task Methods}\\
\midrule
Fine-tuning~\cite{Devlin2019BERT} & 81.42 & 76.15 & 84.50 & 54.17 & 44.45 & 73.28 & 59.64 & 67.66\\
LM-BFF (man)~\cite{Gao2021Making} & 90.75 & 86.60 & 90.50 & 63.62 & 70.77 & 74.05 & 60.27 & 76.65\\
LM-BFF (auto)~\cite{Gao2021Making} & 91.62 & 87.25 & 91.80 & 64.25 & 71.21 & 74.23 & 60.59 & 77.28\\
P-tuning~\cite{Liu2021GPT} & 91.85 & 86.60 & 91.75 & 62.41 & 70.28 & 66.42 & 60.57 & 75.70\\
\midrule
\multicolumn{8}{l}{\em Cross-task Methods w. Learning from Similar Tasks}\\
\midrule
Fine-tuning (mtl)~\cite{Sun2019How} & 83.37 & 79.30 & 84.75 & 41.32 & 48.14 & 53.12 & 59.31 & 64.19\\
Meta Fine-tuing~\cite{Wang2020Meta} & 86.32 & 83.85 & 88.42 & 48.52 & 58.20 & 71.56 & 67.12 & 72.00\\
LM-BFF (mtl)~\cite{Gao2021Making}$^{*}$ & 91.97 & 87.45 & 90.70 & 69.09 & 75.90 & 72.18 & 67.40 & 79.24\\
P-tuning (mtl)~\cite{Liu2021GPT}$^{*}$ & 93.12 & 87.75 & 91.35 & 68.83 & 74.24 & 70.83 & 69.99 & 79.44\\
\textbf{\emph{TransPrompt}}~\cite{wang2021transprompt} & \textbf{93.58} & \bf 88.80 & \bf 92.00 & \bf 71.90 & \textbf{76.99} & \bf 75.98 & \bf 75.80 & \bf 82.15\\
\midrule
\multicolumn{8}{l}{\em Cross-task Methods w. Learning from Distant Tasks}\\
\midrule
Meta-DTL~\cite{Wang2021Meta} & 82.15 & 77.23 & 84.52 &  52.35 & 53.42 & 69.82 & 65.42 & 69.27\\
LM-BFF (mtl)~\cite{Gao2021Making}$^{*}$ & 90.63 & 85.65 & 89.95 & 63.37 & \bf 73.34 & 70.39 & 67.85 & 77.43 \\
P-tuning (mtl)~\cite{Liu2021GPT}$^{*}$ & 91.05 & 86.00 & 90.30 & 63.49 & 72.12 & 71.07 & 70.99 & 77.86 \\
\textbf{\emph{TransPrompt}}~\cite{wang2021transprompt} & 90.47 & 86.05 & 90.30 & 62.56 & 71.97 & 71.83 & 70.17 & 77.62 \\
\bf \emph{\model} & \bf 91.97 & \bf 87.60 & \bf 91.05 & \bf 65.58 & 72.78 & \bf 73.75 & \bf 71.47 & \bf 79.53 \\
\midrule
\end{tabular}
\end{small}
\end{table*}

\begin{table*}
\centering
\caption{\label{tab:ablation}
Ablation study of~\emph{\model} for few-shot learning. Experimental results are reported on the testing sets in terms of accuracy (\%). \textbf{Similar} and \textbf{Distant} denote the similar and distant cross-task learning, respectively.
}
\begin{tabular}{l | c c | c c | c c | c c}

\midrule
\multirow{2}{*}{\textbf{Task}} & \multicolumn{2}{c|}{\bf w/o. Prototype-based De-biasing} & \multicolumn{2}{c|}{\bf w/o. Entropy-based De-biasing} & \multicolumn{2}{c|}{\bf w/o. Both} &
\multicolumn{2}{c}{\bf Full Implementation} \\
\cmidrule{2-9}
 & \textbf{Similar} & \textbf{Distant} & \textbf{Similar} & \textbf{Distant} & \textbf{Similar} & \textbf{Distant} & \textbf{Similar} & \textbf{Distant} \\
\midrule
SST-2 & 92.90 & 90.64 & \bf 93.18 & \bf 91.28 & 92.67 & 90.10 & 93.58 & 91.97\\
MR & 87.97 & 86.30 & \bf 88.14 & \bf 87.30 & 87.75 & 86.05 & 88.80 & 87.60\\
CR & 91.50 & \bf 90.55 & \bf 91.70 & 90.50 & 91.07 & 89.80 & 92.00 & 91.05\\
\midrule
MNLI & 67.72 & 63.77 & \bf 69.74 & \bf 65.05 & 67.08 & 63.86 & 71.90 & 65.58\\
SNLI & \bf 76.76 & 70.29 & 76.66 & \bf 71.15 & 76.08 & 70.91 & 76.99 & 72.78\\
\midrule
MRPC & 75.98 & 72.07 & \bf 76.72 & \bf 73.00 & 68.38 & 65.70 & 75.98 & 73.75\\
QQP & 73.44 & \bf 71.03 & \bf 74.00 & 70.98 & 73.02 & 68.40 & 75.80 & 71.47\\
\midrule
\bf Average & 80.90 & 77.81 & \bf 81.45 & \bf 78.47 & 79.44 & 76.40 & 82.14 & 79.93\\
\midrule
\end{tabular}
\end{table*}

For a fair comparison, we re-produce all baselines based on their open-source codes under the same settings. Our own~\emph{\model} algorithm is implemented in PyTorch and runs with NVIDIA V100 GPUs. In default, we set $\zeta=0.5$, $\gamma=0.001$ and $\lambda_2=0.01$. The parameter regularizers are the same as in~\cite{Liu2021GPT}.
The model is trained with the Adam optimizer~\cite{Kingma2015Adam}. The batch size is set as 6 and 16 in few-shot and full scenarios, respectively. The model architecture of prompt encoders is the same as~\cite{Liu2021GPT}. Therefore, the increased number of parameters of~\emph{\model} remains minimal. 
We further tune the learning rates, epochs, and the length of pseudo tokens $I$, with results reported in the following experiments.

\subsection{General Experimental Results}

The results of~\emph{\model} and all baselines on all seven testing sets for few-shot learning are shown in Table~\ref{tab:few-shot}. 
From the experimental results, we have drawn the following conclusions:
\begin{itemize}
    \item Prompting baselines (such as LM-BFF and P-tuning) outperform the standard fine-tuning method by a large margin in both single-task learning and cross-task learning settings. This shows prompts are useful for few-shot learning. 
    
    \item Based on our re-production, LM-BFF and P-tuning have similar performance. Yet, automatically-mined prompts are still slightly better than manually-compiled prompts for LM-BFF. The gains between LM-BFF and P-tuning over similar and distant cross-task learning are 0.20\% and 0.43\%, respectively, which shows that continuous prompt embeddings are more suitable for multi-task learning than discrete prompts. 
    
    \item Comparing multi-task learning approaches to similar tasks, we find that all methods trained in distant scenarios are decreased over most of the datasets. The results clearly and intuitively reflect the phenomenon that there is a large difference between each type of task. Hence, this setting is significantly more challenging.

    \item As for MMA across distant tasks, the performance of our previous version \emph{TransPrompt}~\cite{wang2021transprompt} is similar to other prompt-based baselines, which indicates that existing prompt-based methods are hard to deal with different types of tasks. Fortunately, we find that the performance can be improved by the utilization of external task type-specific prompting knowledge (i.e.,  task-type descriptions). Specifically, according to the overall averaged results in Table~\ref{tab:few-shot}, \emph{\model} outperforms all baselines, and the average improvement is around 2\% in terms of accuracy, compared to the strongest baseline (i.e., the multi-task version of P-tuning). 
    
    \item For fair comparison in cross-task learning, we also choose two state-of-the-art fine-tuned baselines named MFT and Meta-DTL, which dedicates to meta-learning over similar tasks and distant tasks, respectively. Results show that our method still performs better in both settings.
    
    \item Finally, we have conducted the~\emph{paired t-tests} over the results produced on all tasks. The results show that the improvement of~\emph{\model} is statically significant (with the $p$-value $p<0.01$). This further demonstrates the effectiveness of the proposed~\emph{\model}. 
\end{itemize}



In the following, we study how~\emph{\model} improves the performance in various aspects.



\begin{figure*}[t]
\centering
\begin{tabular}{cc}
\begin{minipage}[t]{0.32\linewidth}
\subfigure[Tuning the learning epoch of the meta-learner during MMA.]{
    \includegraphics[width = 0.95\linewidth]{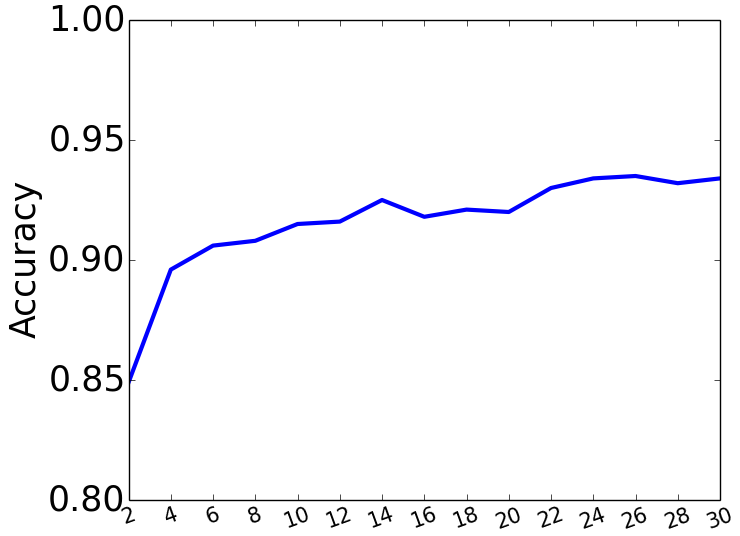}
}
\end{minipage}
\begin{minipage}[t]{0.32\linewidth}
\subfigure[Tuning the learning epoch of the fine-tuned PLM during TMS.]{
    \includegraphics[width = 0.95\linewidth]{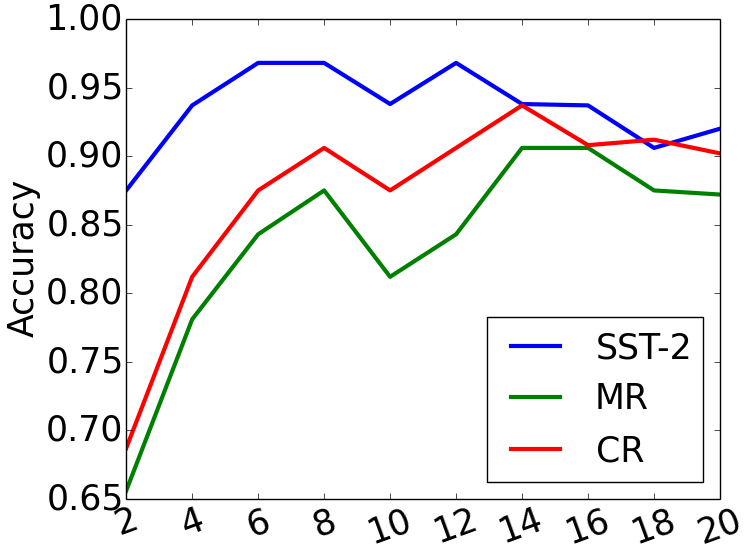}
}
\end{minipage}
\begin{minipage}[t]{0.32\linewidth}
\subfigure[Tuning the learning rate.]{
    \includegraphics[width = 0.95\linewidth]{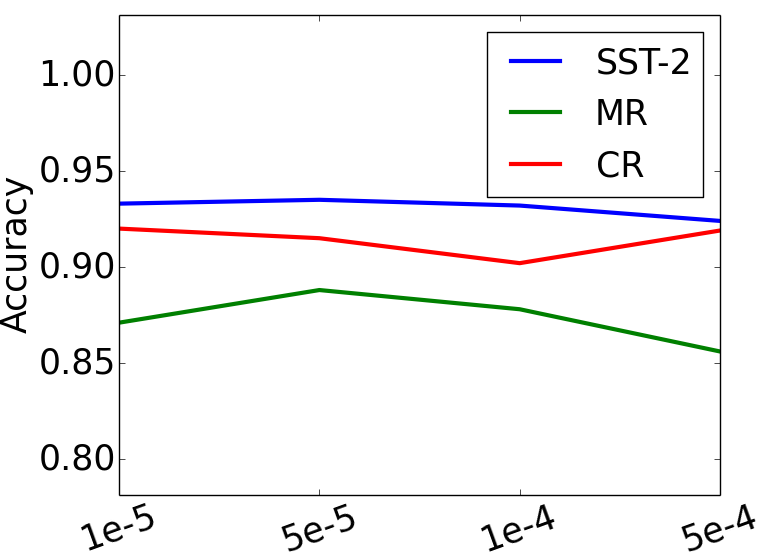}
}
\end{minipage}
\end{tabular}
\caption{Parameter analysis of the learning epoch and the learning rate for learning across similar tasks. The underlying task group is sentiment analysis.}
\label{fig:tune}
\end{figure*}


\begin{table}
\centering
\caption{\label{tab:ablation-desc}
The effectiveness of the task type descriptions and corresponding intra/inter-type prompt embeddings.
}
\begin{tabular}{l | l l l | l}
\midrule
\bf Task & \bf SST-2 & \bf MR & \bf CR & \bf Average\\
\midrule
w/o. description & 90.98 & 86.70 & 90.65 & 89.44 \\
w/o. inter-type & 90.64 & 86.20 & 90.45 & 89.10 \\
w/o. intra-type & 90.64 & 86.15 & 90.50 & 89.10 \\
w/o. all & 90.47 & 86.05 & 90.30 & 88.94 \\
\midrule
\emph{\model} & \bf 91.97 & \bf 87.60 & \bf 91.05 & \bf 90.21 \\
\midrule
\end{tabular}
\end{table}

\subsection{Ablation Studies}

In the~\emph{\model} framework, we propose two de-biasing techniques to improve the effectiveness of the meta-learner, i.e,~\emph{prototype-based} and~\emph{entropy-based}. Here, we remove each one and all de-biasing techniques and implement the variants of~\emph{\model}.
We conduct experiments over both similar and distant tasks. As shown in Table~\ref{tab:ablation},
both de-biasing techniques are proved effective for~\emph{\model}. Particularly,~\emph{prototype-based de-biasing} plays a slightly more important role than~\emph{entropy-based de-biasing} in most scenarios. We conclude that de-biasing the meta-learner is crucial for obtaining cross-task knowledge.

In addition, we also study how \emph{\model} performs well in distant cross-task scenarios. We take SST-2, MR, and CR as examples, and test the performance when removing task type descriptions and corresponding prompt encoders. Results in Table~\ref{tab:ablation-desc} demonstrate that each of techniques is vital for cross-task distant learning.

\begin{table}
\centering
\caption{\label{tab:pseudo_length}
The efficiency of the pseudo prompt length $I$. We conduct experiments on sentiment analysis in few-shot cross-task learning from similar tasks.
}
\begin{tabular}{l | c c c }
\midrule
\bf Task & \bf SST-2 & \bf MR & \bf CR \\
\midrule
$I=1$ & \textbf{93.58} & 88.05 & 91.70 \\
$I=2$ & 93.14 & \textbf{88.25} & \textbf{91.85} \\
$I=3$ & 92.67 & 87.85 & 91.05 \\
$I=4$ & 92.97 & 87.30 & 91.05 \\
\midrule
\end{tabular}
\end{table}

\begin{figure*}[t]
\centering
\begin{tabular}{cc}
\begin{minipage}[t]{0.32\linewidth}
\subfigure[MMA training across $M (M\in\{0,1,2,3\})$ tasks from sentiment analysis, and fine-tuning over MNLI, SNLI, MRPC and QQP.]{
    \includegraphics[width =0.95\linewidth]{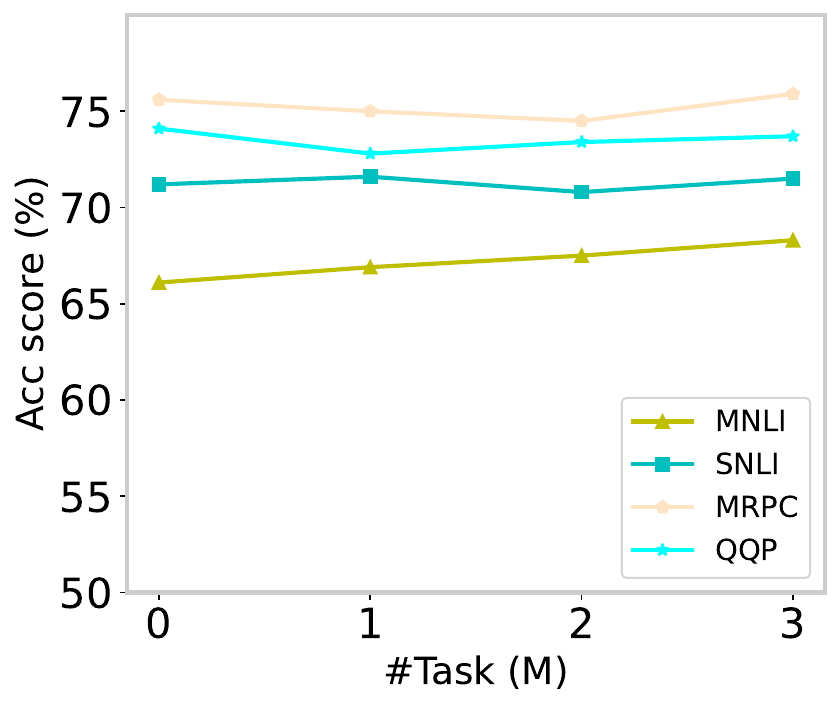}
}
\end{minipage}
\begin{minipage}[t]{0.32\linewidth}
\subfigure[MMA training across $M (M\in\{0,1,2\})$ tasks from NLI, and fine-tuning over SST-2, MR, CR, MRPC and QQP.]{
    \includegraphics[width = 0.95\linewidth]{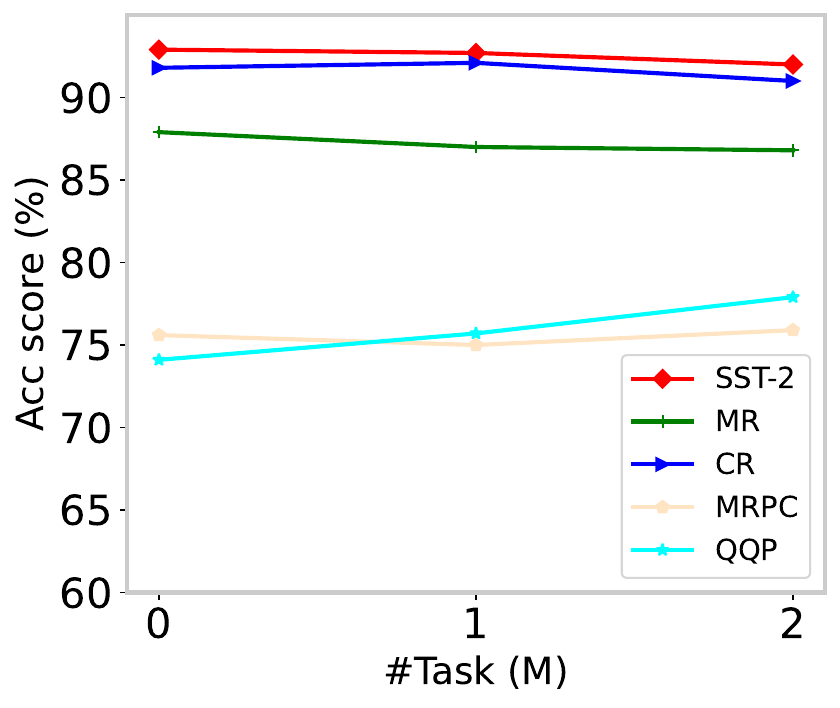}
}
\end{minipage}
\begin{minipage}[t]{0.32\linewidth}
\subfigure[MMA training across $M (M\in\{0,1,2\})$ tasks from paraphrasing, and fine-tuning over SST-2, MR, CR, MNLI and SNLI.]{
    \includegraphics[width = 0.95\linewidth]{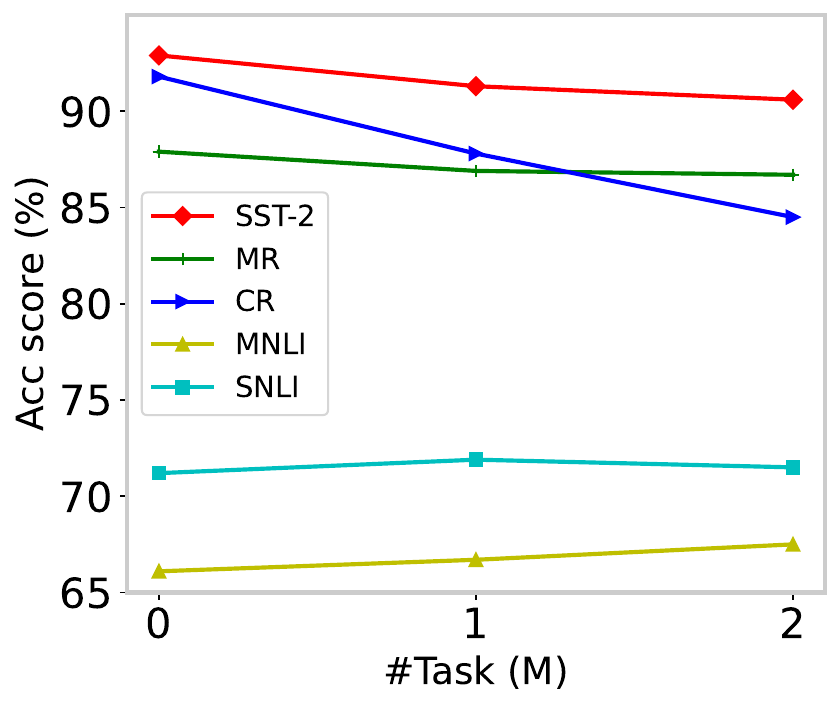}
}
\end{minipage}
\end{tabular}

\caption{Model generalization efficiency in distant cross-task settings. For each figure, the horizontal axis denotes the number $M$ of tasks from the corresponding task group for training the meta-learner $\mathcal{F}_{meta}$, while the vertical axis represents the accuracy of each target unseen distant task $\mathcal{T}^{*}$ in the TMS stage. $M=0$ is a special situation that represents having the TMS stage only.}
\label{fig:generalization_efficiency}
\end{figure*}

\subsection{Parameter Analysis}

We further tune the learning epoch and the learning rate during the training process of~\emph{\model} and report the performance over the development sets. Consider the similar-task learning scenario. We illustrate the results over SST-2, MR, and CR, shown in Figure~\ref{fig:tune}. We fix the learning rate to be 1e-5 and tune the learning epochs. Figure~\ref{fig:tune}(a) shows that the performance of the meta-learner becomes stable over 20 epochs (which is tested on the combination of three development sets for SST-2, MR, and CR). 
Figure~\ref{fig:tune}(b) gives the results of the three tasks during TMS. We can see that \emph{\model} only needs a small number of training steps to reach the best performance on development sets of multiple tasks. It implies that we can leverage an early stopping strategy during experiments. 
For learning rate selection, we fix the training epoch to 20 and tune the learning rate from 1e-5 to 5e-4. As seen in Figure~\ref{fig:tune}(c), the learning rate should be set in the range of 1e-5 to 5e-5.

Except for the analysis of training parameters, we also explore the efficiency of the hyper-parameter in the model, i.e., the length of pseudo prompt tokens. We choose all tasks from sentiment analysis and conduct experiments in few-shot distant cross-task settings. Results in Table~\ref{tab:pseudo_length} suggest that the length of pseudo prompts can influence the model accuracy. Based on our results, we recommend setting $I\in\{1,2\}$.

\begin{table}
\centering
\caption{\label{tab:generalize}
Model generalization results in terms of accuracy (\%). LM-BFF and P-tuning are strong baselines w/o. the usage of the meta-learner.
}
\begin{tabular}{l | l l l | l}
\midrule
\bf Task & \bf SST-2 & \bf MR & \bf CR & \bf Average\\
\midrule
LM-BFF (m) & 90.75 & 86.60 & 90.50 & 89.28\\
LM-BFF (a) & 91.62 & 87.25 & 91.80 & 90.22\\
P-tuning & 91.85 & 86.60 & 91.75 & 90.05\\
\midrule
\emph{\model} & \bf 93.35 & \bf 88.25 & \textbf{91.85} & \bf 91.15\\
\midrule
\end{tabular}
\end{table}

\begin{table*}
\vspace{-.5em}
\centering
\caption{\label{tab:full}
The testing results of~\emph{\model} and baselines with full training sets in terms of accuracy (\%).
}
\begin{small}
\begin{tabular}{l | ccc | cc | cc | c }
\midrule
\multirow{2}{*}{\textbf{Method}} & \multicolumn{3}{c|}{\bf Task: Sentiment Analysis} & \multicolumn{2}{c|}{\bf Task: NLI} & \multicolumn{2}{c|}{\bf Task: paraphrasing} & \multirow{2}{*}{\textbf{Average}}\\
\cmidrule{2-8}
 & \textbf{SST-2} & \textbf{MR} & \textbf{CR} & \textbf{MNLI} & \textbf{SNLI} & \textbf{MRPC} & \textbf{QQP}\\
\midrule
\multicolumn{8}{l}{\em Single-task Baselines}\\
\midrule
Fine-tuning~\cite{Devlin2019BERT} & 93.00 & 90.15 & 90.90 & 82.87 & 87.87 & 72.28 & 89.53 & 86.65\\
LM-BFF (man)~\cite{Gao2021Making} & 93.65 & 88.50 & 90.98 & 87.23 & 91.10 & 88.75 & 85.12 & 89.33\\
LM-BFF (auto)~\cite{Gao2021Making} & 93.81 & 88.75 & 91.25 & 87.01 & 91.51 & \bf 88.97 & 83.12 & 89.20\\
P-tuning~\cite{Liu2021GPT} & 93.69 & 90.10 & 90.25 & 87.17 & 91.67 & \bf 88.97 & 90.87 & 90.38\\
\midrule
\multicolumn{8}{l}{\em Cross-task Baselines w. Learning from Similar Tasks}\\
\midrule
Fine-tuning (mtl)~\cite{Sun2019How} & 94.72 & 90.65 & 91.05 & 87.10 & 91.80 & 69.85 & 90.20 & 87.91\\
Meta Fine-tuing~\cite{Wang2020Meta} & 95.70 & 91.25 & 91.42 & 83.67 & 89.48 & 78.92 & 89.72 & 88.59\\
LM-BFF (mtl)~\cite{Gao2021Making}$^{*}$ & 95.41 & 90.45 & 91.50 & 86.76 & 88.25 & 69.36 & 90.32 & 87.43\\
P-tuning (mtl)~\cite{Liu2021GPT}$^{*}$ & 95.30 & 90.40 & 90.08 & 86.97 & 91.48 & 68.87 & 90.59 & 87.67\\
\textbf{\emph{TransPrompt}}~\cite{wang2021transprompt} & \bf 96.05 & \bf 91.78 & \bf 91.59 & \bf 88.70 & \bf 91.88 & 86.87 & \bf 91.27 & \bf 91.16\\
\midrule
\multicolumn{8}{l}{\em Cross-task Methods w. Learning from Distant Tasks}\\
\midrule
Meta-DTL~\cite{Wang2021Meta} &  93.70 & 90.20 & 90.14 & 82.65 & 88.10 & 74.86 & 88.84 & 86.92 \\
LM-BFF (mtl)~\cite{Gao2021Making}$^{*}$ & 94.10 & 90.65 & 91.30 & 86.98 & 90.10 & 81.07 & 88.83 & 89.00 \\
P-tuning (mtl)~\cite{Liu2021GPT}$^{*}$ & 95.04 & 90.35 & 91.10 & 88.05 & 90.65 & 82.72 & 90.16 & 89.72 \\
\textbf{\emph{TransPrompt}}~\cite{wang2021transprompt} & 94.61 & 90.80 & 91.25 & 87.93 & 91.62 & 85.90 & 91.47 & 90.57 \\
\bf \emph{\model} & \bf 95.53 & \bf 91.15 & \bf 91.95 & \bf 88.49 & \bf 92.17 & \bf 86.30 & \bf 91.53 & \bf 91.10\\
\midrule
\end{tabular}
\end{small}
\end{table*}

\subsection{Model Generalization to New Unseen Tasks}
One advantage of~\emph{\model} is that it can train a meta-learner with~\emph{cross-task transferable knowledge} encoded. 
In this set of experiments, we consider two scenarios including generalization from similar tasks and distant tasks. All experiments are in few-shot settings.

For the \emph{adaptation} from similar tasks, we consider three tasks of sentiment analysis: SST-2, MR, and CR. Each time, we train the meta-leaner over two out of the three datasets (the MMA stage) and then make the model generalized to each of the other tasks (the TMS stage). In these settings, there are no training examples that overlap between the two stages. For example, we train the meta-learner $\mathcal{F}_{meta}$ over the few-shot SST-2 and MR datasets and then take the TMS step over the few-shot CR dataset. Here, the meta-learner has no knowledge of CR before the TMS step. So, it can validate whether the model can solve new unseen tasks with knowledge of similar tasks.
We test whether the usage of the meta-learner is better than simply applying LM-BFF~\cite{Gao2021Making} or P-tuning~\cite{Liu2021GPT} initialized from PLMs. From Table~\ref{tab:generalize}, it is clearly reflected that the meta-learner brings improvements in all three cases, hence generalizing to new tasks accurately.

For the \emph{generalization} from distant tasks, we explore how many distant tasks can make contributions to each target task. For example, we choose one group of tasks (e.g., SST-2, MR, and CR in sentiment analysis.), and randomly select $M$ tasks from this group for training the meta-learner $\mathcal{F}_{meta}$. Then, we make the model generalized to each of the other distant tasks (i.e., MNLI, SNLI, MRPC, and QQP). In this setting, the model can not see any data from the target task during MMA, which can be better to test the effectiveness of model generalization.
As shown in Figure~\ref{fig:generalization_efficiency}, we draw some conclusions:
\begin{itemize}
    \item The overall results show that our method can fuse semantics knowledge from distant tasks to enhance the effectiveness of most datasets.
    
    \item Tasks from sentiment analysis have a slight improvement for NLI and paraphrasing, while the improvement is not obvious. As shown in Figure~\ref{fig:generalization_efficiency}(a), when meta-training overall sentiment tasks, only the MNLI task increases more than 1.0\%.
    
    \item It is difficult for the task form sentiment analysis to capture transferable knowledge from tasks of NLI and paraphrasing. Especially, when we train $\mathcal{F}_{meta}$ over tasks from paraphrasing, the performance of sentiment analysis decreases. We guess that the semantics of sentence-pair tasks are highly different from single-sentence tasks, which improves the difficulty of knowledge transfer.
    
    \item Tasks from NLI and paraphrasing can be enhanced from each other. As shown in Figure~\ref{fig:generalization_efficiency}(b) and~\ref{fig:generalization_efficiency}(c), with the increase of training data, there is a slight improvement in the NLI and paraphrasing tasks.
\end{itemize}

\subsection{Learning with Full Datasets}

Besides, we are also interested in how~\emph{\model} performs when learning with full training sets. We follow the base-scale experimental settings in~\cite{Liu2021GPT}, with the RoBERTa base model (with 109M parameters) as the underlying PLM. 
The results are presented in Table~\ref{tab:full}.
On average, \emph{\model} outperforms all baselines by around 1\% to 5\% regardless of whether the model is replaced in single-task or cross-task scenarios.
This shows that our proposed paradigm can be of help in non-few-shot learning scenarios by learning from similar or distant tasks.

Another interesting finding is that when it comes to multi-task learning, the performance of LM-BFF~\cite{Gao2021Making} and P-tuning~\cite{Liu2021GPT} drops, compared to the single-task setting. A most possible cause is that with a large amount of training data from other tasks, existing prompt-based approaches may capture non-transferable knowledge that is harmful to the target task.
In contrast, the two-step paradigm of~\emph{\model} learns different types of knowledge at different steps (i.e., the universal knowledge in MMA, and the task-specific knowledge in TMS), hence producing better results. Overall,~\emph{\model} is competitive in standard fine-tuning scenarios with datasets from similar NLP tasks available.

\subsection{Case Studies}

For a more initiative understanding of which instances are~\emph{more transferable across tasks}, in Table~\ref{tab:case}, several review texts from SST-2, MR, and CR with high and low prototype scores are presented.
Although these texts come from different tasks, our algorithm is able to find texts that express general polarities instead of specific points. For instance, ``time waster'', ``remarkable'' and ``5 stars'' are strong indicators of polarities, which receive high scores generated. 
In contrast, review texts with low scores are overly specific and hence are less transferable across tasks.
Hence, our meta-learner truly captures the transferable knowledge.

\begin{table*}
\centering
\caption{Cases of review texts in SST-2, MR and CR with high and low cross-task prototype scores.} 
\begin{small} 
\begin{tabular}{l | l | l | l}  
\midrule
\bf Score & \bf Task & \bf Review Text & \bf Label\\
\midrule
& SST-2 & There are many definitions of ``time waster'' but this movie must surely be one of them... & NEG\\
High & MR & It is most remarkable not because of its epic scope, but because of the startling intimacy... & POS\\
& CR & 5 stars all the way! & POS\\
\midrule
& SST-2 & It's a treat watching show, a British stage icon, melting under the heat of phocion's attentions.  & POS\\
Low & MR & Humorous, artsy, and even cute, in an off-kilter, dark, vaguely disturbing way.  & POS\\
& CR & However, the calls constantly drop in my area and I experience mega-static, to the point... & NEG\\
\midrule
\end{tabular} 
\end{small}
\label{tab:case} 
\end{table*}

\begin{figure}
\centering
\includegraphics[width=0.7\linewidth]{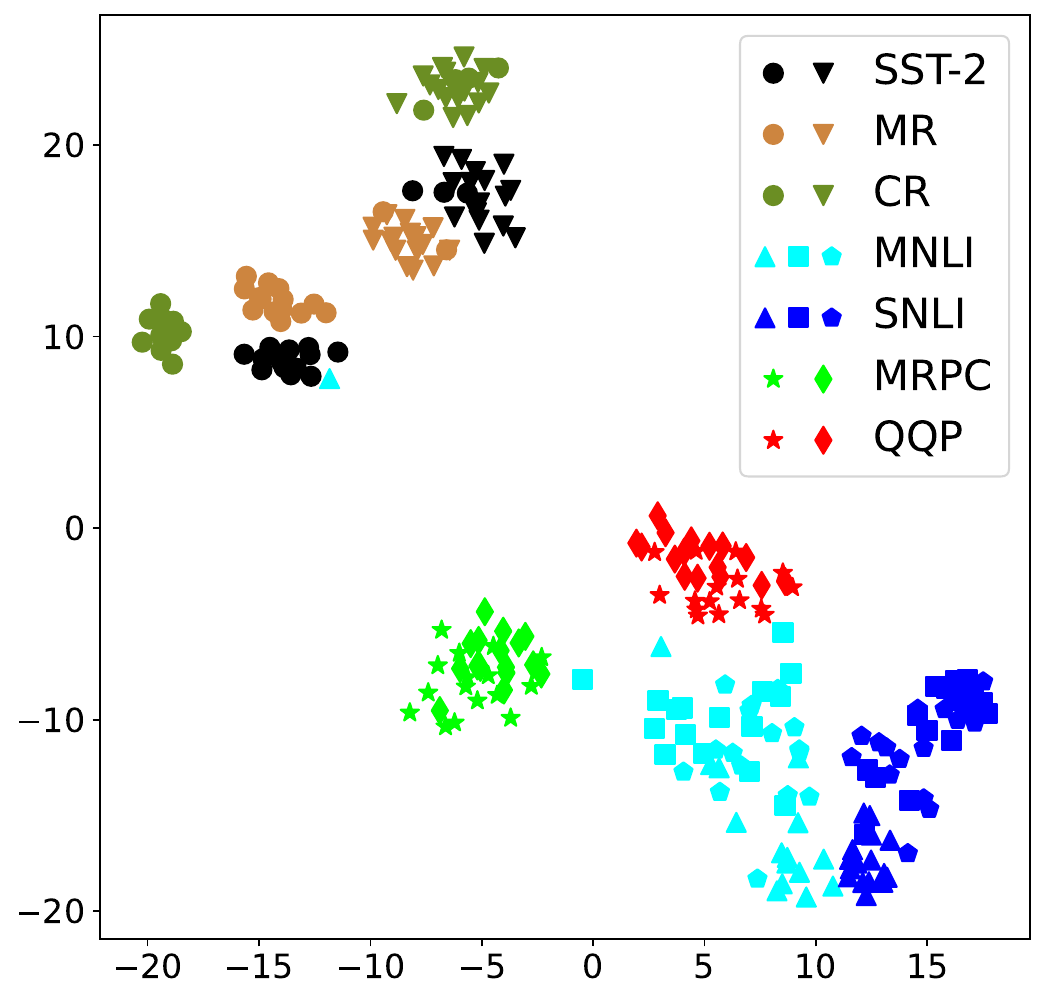}
\caption{Visualizations of sentence embeddings of few-shot scenarios over all datasets. Different colors represent different datasets. Different shapes represent different categories. (Best viewed in color.)}
\label{fig:visual}
\end{figure}

\subsection{Visualizations}

Furthermore, we visualize one set of sentences from all development sets in the few-shot settings to gain more insight into the model performance. Specifically, we directly aggregate all few-shot training data from all datasets (i.e., 256 examples) to train the meta-learner, and then perform inference over corresponding few-shot development sets. For each sentence, we choose the representations at the position of \texttt{[MASK]} as the sentence embeddings. It can be seen from Figure~\ref{fig:visual} that our proposed technique plays an important role in sentence representations. Intuitively, we find that the tasks from sentiment analysis (which are single-sentence tasks) are more similar to each other, while NLI and paraphrasing (which are sentence-pair tasks) have similar distributions. However, there exists a large difference between single-sentence and sentence-pair tasks, which makes the transfer across distant tasks challenging. In addition, the results also demonstrate that the classification performance of sentiment analysis is better than others.

\section{Conclusion}
\label{sec:con}

In this paper, we present the~\emph{\model} framework for few-shot learning across similar and distant text classification tasks based on continuous prompt embeddings and task type description.
Experimental results show that \emph{\model} consistently outperforms strong baselines in both few-shot learning and standard fine-tuning settings. Additionally, we find that the meta-learner trained by~\emph{\model} can be adapted to previously unseen tasks easily. 
Future works include i) expanding our work to other multiple NLP tasks 
during cross-task learning. ii) exploring how~\emph{\model} is applied to other PLMs apart from BERT-style models. iii) devoting to the knowledge-enhanced prompt-based fine-tuning for distant cross-task learning.


\begin{IEEEbiographynophoto}{Jianing Wang}
is currently a Ph.D student on the School of Data Science and Engineering (DaSE), East China Normal University (ECNU), China,
focusing on deep learning and natural language processing. 
His research interests include pre-trained language models, few-shot learning, and knowledge graph. He has published several papers in major conferences and journals.
\end{IEEEbiographynophoto}

\begin{IEEEbiographynophoto}{Chengyu Wang}
received his Ph.D degree from East China Normal University (ECNU) in 2020. Currently, he works on deep learning algorithms on various topics for Alibaba Cloud Platform of AI (PAI)  including natural language processing, human speech understanding, transfer learning, and few-shot learning. He has published 50+ research papers in top-tier international conferences and journals, such as ACL, KDD, SIGIR, WWW, AAAI, TKDE, CIKM, EMNLP, etc.
\end{IEEEbiographynophoto}

\begin{IEEEbiographynophoto}{Cen Chen} 
is currently an associate professor at the School of Data Science and Engineering (DaSE), East
China Normal University (ECNU), China. She received a Ph.D. degree from the School of Information Systems, Singapore Management University in 2017. She was a visiting scholar at Carnegie Mellon University from 2015 to 2016. She has published papers in top data mining and artificial intelligence conferences and journals including WWW, AAAI, IJCAI, SIGIR, and VLDB. Her current research interests include privacy-preserving machine learning and NLP applications.
\end{IEEEbiographynophoto}

\begin{IEEEbiographynophoto}{Ming Gao}
received the doctorate degree from School of Computer Science, Fudan University, Shanghai, China. He is currently a full professor on School of Data Science and Engineering (DASE), East China Normal University (ECNU), China. His research interests include knowledge engineering, user profiling, social network analysis, and mining. His works appear in major international journals and conferences, including Data Mining and Knowledge Discovery, IEEE Transactions on Knowledge and Data Engineering, Knowledge and Information Systems, ICDE, ICDM, SIGIR, etc.
\end{IEEEbiographynophoto}

\begin{IEEEbiographynophoto}{Jun Huang}
received a Ph.D. degree in Modern Physics from the University of Science and Technology of China in 2008. He was an associate research fellow of China Academy of Engineering Physics. Now he leads a team for developing AI algorithms on the Platform of AI of Alibaba Group, responsible for developing innovative algorithms and platforms such as deep learning, transfer learning, federal learning, etc. to serve important internal and external business of Alibaba. His interests include high-performance distributed implementation of AI algorithms and applying them to real applications.
\end{IEEEbiographynophoto}

\begin{IEEEbiographynophoto}{Aoying Zhou}
is a professor in the School of Data Science and Engineering (DaSE), East China Normal University (ECNU), China, where he is heading DaSE and vice president of ECNU. He is the winner of the National Science Fund for Distinguished Young Scholars supported by NSFC and the professorship appointment under Changjiang Scholars Program of Ministry of Education. His research interests include Web data management, data intensive computing, in-memory cluster computing, and benchmark for big data. His works appear in major international journals and conferences, including IEEE Transactions on Knowledge and Data Engineering, Proceedings of the VLDB Endowment, SIGMOD, SIGIR, KDD, WWW, ICDE, and ICDM, etc.
\end{IEEEbiographynophoto}

\vfill

\end{document}